# Deep Learning Model for Amyloidogenicity Prediction using a Pre-trained Protein LLM

Zohra YAGOUB* and Hafida BOUZIANE

Université des Sciences et de la Technologie d'Oran Mohamed Boudiaf, USTO-MB,
BP 1505, El M'Naouer, 31000 Oran, Algeria
*zohra.yagoub@univ-usto.dz

**Abstract.** The prediction of amyloidogenicity in peptides and proteins remains a focal point of ongoing bioinformatics. The crucial step in this field is to apply advanced computational methodologies. Many recent approaches to predicting amyloidogenicity within proteins are highly based on evolutionary motifs and the individual properties of amino acids. It is becoming increasingly evident that the sequence information-based features show high predictive performance. Consequently, our study evaluated the contextual features of protein sequences obtained from a pretrained protein large language model leveraging bidirectional LSTM and GRU to predict amyloidogenic regions in peptide and protein sequences. Our method achieved an accuracy of 84.5% on 10-fold cross-validation and an accuracy of 83% in the test dataset. Our results demonstrate competitive performance, highlighting the potential of LLMs in enhancing the accuracy of amyloid prediction.

## 1 Introduction

The primary structure of a protein determines its unique linear sequence of amino acids, which shape its three-dimensional conformation. Consequently, the sequence has a crucial role in determining the protein structure and function.

Numerous laboratory experiments have shown that the conditions under which peptides and proteins aggregate are strongly influenced by their amino acids sequences. Specifically, altering the protein sequence through mutations, even by changing a single amino acid, or subjecting the protein to conditions that partially unfold it, can significantly impact its tendency to form amyloid fibrils [1–5]. This insight has been central to advancing protein aggregation propensity prediction by effectively exploiting evolutionary and sequential features encoded within protein sequences by leveraging different machine learning models[6,7]. Among these informations, we cite profile-based features, n-gram features, Position Specific Scoring Matrix and sequence-based features including amino acid composition, transition, and distribution.

Although many computation-based methods have been proposed for the prediction of amyloid proteins, achieving high predictive performance remain challenging issue. In this study, we analyzed the peptide and protein sequences as text to learn patterns and relationships within their amino acids using a pre-trained protein LLM.

## 2 Methods

### 2.1 Datasets

6-residue fragments are often considered the most common and significant contributors to amyloid formation, acting as crucial "hot spots" for initiating amyloid formation[8]. Therefore, we trained our method using the WaltzDB 2.0[1] [9] dataset, consisting of 515 amyloidogenic hexapeptides and 901 non-amyloidogenic hexapeptides. The dataset was randomly split into two datasets: 80% for training and cross-validation and 20% as a test set to evaluate the ability of the proposed LLM to classify hexapeptide sequences.

For evaluating the performance of the proposed method in distinguishing between amyloidogenic and non-amyloidogenic peptides, we exploited the same dataset used in ReRF-Pred[10], consisting of 251 amyloidogenic peptides with different lengths(Pep-251).

The dataset used to validate the effectiveness of our method in identifying amyloidogenic regions in whole protein sequences was AmyPro27 as employed in AggreProt [11]. It consists of 27 sequences of amyloid proteins extracted from AmyPro[2] database with their aggregation-prone region annotated containing less than 50 amino acids.

### 2.2 Hexapeptide sequences features representation

To comprehensively evaluate the performance of the representations generated by the large language models (LLMs) in predicting amyloidogenicity in peptides and proteins, we adopted the pre-trained model ESM-2[12]. It is built upon the Transformer architecture using the BERT encoder. ESM-2 is trained on protein sequences from the UniRef database[13] using a "masked language modeling" technique. Through this, ESM-2 learns intricate patterns and contacts between amino acids, effectively capturing contextual information encoded within the protein sequence. ESM-2 has shown remarkable success in various protein-related tasks, including protein structure prediction and protein function prediction.

In this present work, 1280 per-sequence embeddings are directly generated via averaging embeddings for each amino acid in the input sequence from the ESM-2 model pre-trained with 650 million parameters with the number of layers being 33, representing their numerical internal contact representations for the 1426 WaltzDB hexapeptides.

---

[1] http://waltzdb.switchlab.org/
[2] http://www.amypro.net/

### 2.3 Model Construction

In this study, we investigated and evaluated the performance of two deep learning models using the embeddings obtained to discriminate between amyloidogenic and non-amyloidogenic hexapeptides, as shown in Fig. 1. These embeddings serve as input for bidirectional long-term memory (Bi-LSTM) and bidirectional gated recurrent unit (Bi-GRU), enabling the model to process contextual sequence information in both forward and backward directions. To prevent over-fitting during training, dropout layers are added after each neural network model, allowing robust feature extraction by dropping out some neurons. A final dense layer is used by including a single neuron with sigmoid activation in the output layer to perform binary classification. This hybrid model aims to improve prediction accuracy by integrating LLM embeddings with advanced neural network architectures.

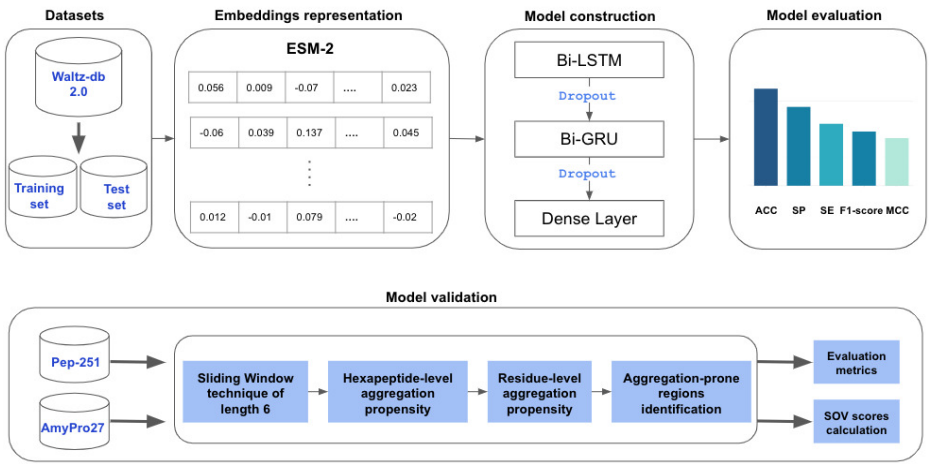

**Fig. 1.** The frame chart of the proposed method.

## 3 Results and Discussion

### 3.1 Ability of the LLMs in classifying hexapeptide sequences

Before testing the effectiveness of our method in distinguishing between amyloid and non-amyloid peptides and proteins. We tested nine machine-learning models using the test dataset to evaluate the proposed LLM ability to identify hexapeptide sequences without integrating additional features. Accuracy, Sensitivity, Specificity, F1-score and Matthew's correlation coefficient(MCC) are used to comprehensively evaluate the prediction performance of these various models, These measures are defined as follows:

$$Accuracy = \frac{TN+TP}{TN+FP+TP+FN} \quad (1)$$

$$Sensitivity = \frac{TP}{TP+FN} \quad (2)$$

$$Specificity = \frac{TN}{TN+FP} \quad (3)$$

$$F1-score = \frac{2TP}{2TP+FP+FN} \quad (4)$$

$$MCC = \frac{TP*TN - FP*FN}{\sqrt{(TP+FP)(TP+FN)(TN+FP)(TN+FN)}} \quad (5)$$

where TP, FP, TN, and FN represent the number of true positive, false positive, true negative, and false negative, respectively.

We can see in Table 1 that accuracy values range from 80% to 83%. While sensitivity and specificity values vary, our model shows the optimum balance achieving 76% and 87%, respectively. The MCC ranges from 0.57 to 0.64, while the F1-scores range from 0.73 to 0.77.

We also evaluated the predictive performance of the proposed deep learning model by performing ten-fold cross-validation, achieving an accuracy of 84.5% and a balance between sensitivity and specificity of 74.5% and 89.7%, respectively.

These findings demonstrate that deep learning and machine learning techniques provide balanced predictions for amyloidogenic hexapeptides, highlighting their ability to learn from the LLMs representations.

**Table 1.** Comparison of different ML models in hexapeptides identification

|                 | Accuracy | Sensitivity | Specificity | F1-score | MCC  |
|-----------------|----------|-------------|-------------|----------|------|
| Adaboost        | 0.80     | 0.74        | 0.83        | 0.73     | 0.57 |
| LGBM            | 0.83     | 0.72        | 0.90        | 0.76     | 0.63 |
| RF              | 0.83     | 0.71        | 0.90        | 0.75     | 0.62 |
| MLP             | 0.80     | 0.73        | 0.85        | 0.73     | 0.58 |
| KNN             | 0.81     | 0.80        | 0.81        | 0.76     | 0.61 |
| LR              | 0.80     | 0.70        | 0.87        | 0.73     | 0.58 |
| ExtraTrees      | 0.83     | 0.72        | 0.90        | 0.76     | 0.64 |
| SVM             | 0.81     | 0.72        | 0.87        | 0.74     | 0.60 |
| Bi-LSTM + Bi-GRU | 0.83    | 0.76        | 0.87        | 0.77     | 0.64 |

### 3.2 Performance of the model in amyloid peptide identification

In this section, we compared our proposed model with state-of-the-art methods by evaluating their ability to discriminate between amyloidogenic and non-amyloidogenic peptides in the Pep-251 dataset. To classify a peptide, we considered it amyloidogenic if the model identified at least one amyloid-prone fragment within its sequence. Initially, the input sequence is segmented into overlapping hexapeptides. Following hexapeptide-level predictions, the results are distributed back to individual amino acids using a sliding window approach and then averaged. The proposed method is compared with six methods, ANuPP[14], ReRF-Pred[10], AmyloGram[15], APPNN[16] and PASTA 2.0[17]. As can be shown in Table 2, Our method demonstrates competitive performance when compared to state-of-the-art amyloid prediction tools. Achieving an accuracy of 0.808, it approximates ReRF-Pred and AmyloGram. Notably, our method exhibits a balanced sensitivity of 0.734, matching ANupp's and ReRF-Pred's. In terms of specificity, our method scores 0.843, outperforming also ANupp, ReRF-Pred, AmyloGram, and APPNN. PASTA 2.0 exhibits a much lower sensitivity even if it has higher specificity. With a Matthews Correlation Coefficient of 0.566, our approach demonstrates a strong correlation between predicted and observed values. This suggests that our method provides a more balanced prediction, effectively distinguishing between amyloidogenic and non-amyloidogenic peptide sequences.

**Table 2.** Performance of our model and other methods in peptides prediction.

|  | Accuracy | Sensitivity | Specificity | MCC |
|---|---|---|---|---|
| Our method | 0.808 | 0.734 | 0.843 | 0.566 |
| ANuPP | 0.733 | 0.734 | 0.732 | 0.440 |
| ReRF-Pred | 0.801 | 0.734 | 0.831 | 0.552 |
| AmyloGram | 0.784 | 0.822 | 0.767 | 0.555 |
| APPNN | 0.769 | 0.848 | 0.733 | 0.542 |
| PASTA 2.0 | 0.833 | 0.506 | 0.983 | 0.603 |

### 3.3 Performance of the model in aggregation-prone regions identification in whole proteins

The effectiveness of the proposed model is validated by the same procedure used for peptide identification on 27 full-length amyloid proteins annotated with their amyloidogenic regions by comparing it with recent and relevant existing methods: AggreProt[11], CrossBeta-RF[18], ANuPP[14], Waltz[19], and FoldAmyloid[20] via Segment OVerlap (SOV) metrics[21] that represents the average overlap between the observed and predicted regions. We calculated SOV for aggregation-prone regions (APR) and non-APR, and average SOV as reported in Table 3. In addition, the number of proteins predicted as non-amyloid was calculated. Our method demonstrated a balanced performance, achieving SOV scores of 53.1 for APR and 54.14 for non-APR, resulting in an average SOV of 53.6. However, it identified only one protein as non-amyloid. AggreProt showed a high SOV for APR, but for non-APR, the SOV was much lower. Despite predicting all proteins as amyloid proteins, CrossBeta-RF had the lowest average SOV of 36.75, with a significant difference between SOV APR (49.0) and SOV non-APR (24.4). On the other hand ANuPP showed a higher average SOV (56.7) but predicted four proteins as non-amyloid, indicating a strong bias towards distinguishing between amyloid and non-amyloid proteins. Waltz displayed a large disparity between APR and non-APR SOV scores of 26.70 and 61.0, respectively. Like ANuPP, it also predicts four proteins as non-amyloid. FoldAmyloid showed comparable scores of 48.3 for APR and 50.0 for non-APR.

In summary, our method offers a balanced prediction across both APR and non-APR segments without significant bias compared to the other predictors.

**Table 3.** Comparison of performances of different APR predictors on AmyPro dataset.

|  | $SOV_{APR}$ | $SOV_{Non-APR}$ | $SOV_{Average}$ | Predicted as non- amyloid |
|---|---|---|---|---|
| Our method | 53.1 | 54.1 | 53.6 | 1 |
| AggreProt | 54.8 | 41.7 | 48.2 | 0 |
| CrossBeta-RF | 49.0 | 24.4 | 36.7 | 0 |
| ANuPP | 53.1 | 60.3 | 56.7 | 4 |
| Waltz | 26.7 | 61.0 | 43.8 | 4 |
| FoldAmyloid | 48.3 | 50.0 | 49.2 | 1 |

## 4 Conclusion

This work shows that amyloidogenic regions may be accurately identified by using contextual features extracted from a protein large language model (ESM-2). By employing bidirectional LSTM and GRU architectures, the experimental results show that our method outperforms state-of-the-art methods. Notably, our approach balances sensitivity and specificity, indicating an accurate and robust ability to distinguish between amyloidogenic and non-amyloidogenic peptide and protein sequences. Future work will focus on evaluating the performance of our model using other pre-trained protein LLMs and integrating additional sequence- and structure-based features to improve the predictive performance.

## References

1. Chiti, F.: Mutational analysis of the propensity for amyloid formation by a globular protein. The EMBO Journal 19(7), 1441–1449 (2000)
2. Ow, S.Y., Dunstan, D.E.: A brief overview of amyloids and Alzheimer's disease: Amyloids and Alzheimer's Disease. Protein Science 23(10), 1315–1331 (2014)
3. Ramírez-Alvarado, M., Merkel, J.S., Regan, L.: A systematic exploration of the influence of the protein stability on amyloid fibril formation in vitro. Proceedings of the National Academy of Sciences of the United States of America 97(16), 8979–8984 (2000)
4. Chiti, F., Calamai, M., Taddei, N., Stefani, M., Ramponi, G., Dobson, C.M.: Studies of the aggregation of mutant proteins *in vitro* provide insights into the genetics of amyloid diseases. Proceedings of the National Academy of Sciences 99(suppl_4), 16419–16426 (2002)
5. Tjernberg, L., Hosia, W., Bark, N., Thyberg, J., Johansson, J.: Charge Attraction and β Propensity Are Necessary for Amyloid Fibril Formation from Tetrapeptides. Journal of Biological Chemistry 277(45), 43243–43246 (2002)
6. Akbar, S., Ali, H., Ahmad, A., Sarker, M.R., Saeed, A., Salwana, E., Gul, S., Khan, A., Ali, F.: Prediction of Amyloid Proteins Using Embedded Evolutionary & Ensemble Feature Selection Based Descriptors With eXtreme Gradient Boosting Model. IEEE Access 11, 39024–39036 (2023)
7. Yang, R., Liu, J., Zhang, Q., Zhang, L.: Multi-view feature fusion and density-based minority over-sampling technique for amyloid protein prediction under imbalanced data. Applied Soft Computing 150, 111100 (2023)
8. Fernandez-Escamilla, A.M., Rousseau, F., Schymkowitz, J., Serrano, L.: Prediction of sequence-dependent and mutational effects on the aggregation of peptides and proteins. Nature Biotechnology 22(10), 1302–1306 (2004)
9. Louros, N., Konstantoulea, K., De Vleeschouwer, M., Ramakers, M., Schymkowitz, J., Rousseau, F.: WALTZ-DB 2.0: an updated database containing structural information of experimentally determined amyloid-forming peptides. Nucleic Acids Research 48(D1), D389–D393 (2020)
10. Teng, Z., Zhang, Z., Tian, Z., Li, Y., Wang, G.: ReRF-Pred: predicting amyloidogenic regions of proteins based on their pseudo amino acid composition and tripeptide composition. BMC Bioinformatics 22(1), 545 (2021)
11. Planas-Iglesias, J., Borko, S., Swiatkowski, J., Elias, M., Havlasek, M., Salamon, O., Grakova, E., Kunka, A., Martinovic, T., Damborsky, J., Martinovic, J., Bednar, D.: AggreProt: a web server for predicting and engineering aggregation prone regions in proteins. Nucleic Acids Research 52(W1), W159–W169 (2024)
12. Lin, Z., Akin, H., Rao, R., Hie, B., Zhu, Z., Lu, W., Smetanin, N., Verkuil, R., Kabeli, O., Shmueli, Y., et al.: Evolutionary-scale prediction of atomic-level protein structure with a language model. Science 379(6637), 1123–1130 (2023)
13. Suzek, B.E., Wang, Y., Huang, H., McGarvey, P.B., Wu, C.H., the UniProt Consortium: UniRef clusters: a comprehensive and scalable alternative for improving sequence similarity searches. Bioinformatics 31(6), 926–932 (2015)
14. Prabakaran, R., Rawat, P., Kumar, S., Michael Gromiha, M.: ANuPP: A Versatile Tool to Predict Aggregation Nucleating Regions in Peptides and Proteins. Journal of Molecular Biology 433(11), 166707 (2021)
15. Burdukiewicz, M., Sobczyk, P., Rödiger, S., Duda-Madej, A., Mackiewicz, P., Kotulska, M.: Amyloidogenic motifs revealed by n-gram analysis. Scientific Reports 7(1), 12961 (2017)
16. Família, C., Dennison, S.R., Quintas, A., Phoenix, D.A.: Prediction of Peptide and Protein Propensity for Amyloid Formation. PLOS ONE 10(8), e0134679 (2015)
17. Walsh, I., Seno, F., Tosatto, S.C., Trovato, A.: PASTA 2.0: an improved server for protein aggregation prediction. Nucleic Acids Research 42(W1), W301–W307 (2014)
18. Kajava, A.: Developing machine-learning-based amyloid predictors with Cross-Beta DB. bioRxiv 21(2), e14510 (2024)

19. Maurer-Stroh, S., Debulpaep, M., Kuemmerer, N., De La Paz, M.L., Martins, I.C., Reumers, J., Morris, K.L., Copland, A., Serpell, L., Serrano, L., Schymkowitz, J.W.H., Rousseau, F.: Exploring the sequence determinants of amyloid structure using position-specific scoring matrices. Nature Methods 7(3), 237–242 (2010)
20. Garbuzynskiy, S.O., Lobanov, M.Y., Galzitskaya, O.V.: FoldAmyloid: a method of prediction of amyloidogenic regions from protein sequence. Bioinformatics 26(3), 326–332 (2010)
21. Zemla, A., Venclovas, E., Fidelis, K., Rost, B.: A modified definition of Sov, a segment-based measure for protein secondary structure prediction assessment. Proteins: Structure, Function, and Genetics 34(2), 220–223 (1999)